\documentclass[runningheads]{llncs}

\usepackage[mobile]{eccv}

\usepackage{eccvabbrv}

\usepackage{graphicx}
\usepackage{booktabs}
\usepackage{multirow}
\usepackage[table,xcdraw]{xcolor}
\usepackage{amsmath}
\usepackage{pifont}
\usepackage{tabularx} 
\usepackage{float}
\usepackage{stfloats}

\usepackage[accsupp]{axessibility}  

\usepackage{hyperref}

\usepackage{orcidlink}

\begin{document}

\title{GDPO-Listener: Expressive Interactive Head Generation via Auto-Regressive Flow Matching and Group reward-Decoupled Policy Optimization}

\titlerunning{GDPO-Listener}

\author{Zhangyu Jin\inst{1}, Maksim Siniukov\inst{1}, Deuksin Kwon\inst{1}, Ashutosh Chaubey\inst{1}, Mohammad Soleymani\inst{1}}

\authorrunning{Z.~Jin et al.}

\institute{University of Southern California, Institute for Creative Technologies\\
\email{\{zjin,msiniukov,dkwon,achaubey,soleymani\}@ict.usc.edu}\\[1em]
\url{https://jinzhangyu.github.io/projects/GDPO-Listener/}
\vspace{-1em}
}

\maketitle

\begin{figure}[h]
  \centering
  \includegraphics[height=4cm]{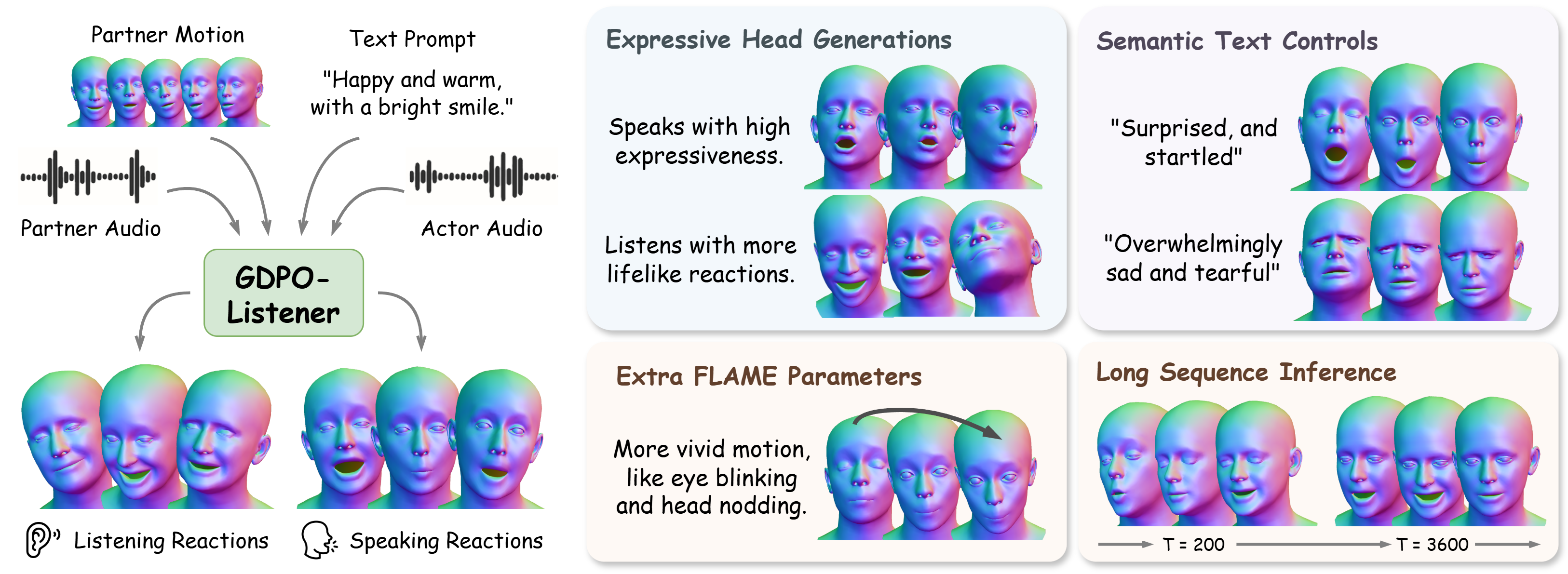}
  \captionsetup{width=11cm}
  \caption{\textbf{GDPO-Listener}. Our framework generates expressive speaking and listening head reactions from multimodal dyadic conversational inputs. By utilizing an expanded FLAME parameter space, it naturally supports eye blinking and head nodding. Furthermore, it enables explicit semantic text control to ensure contextually appropriate responses and maintains stable dynamics during long-sequence inference.
  }
  \label{fig:teaser}
  \vspace{-2em}
\end{figure}

\begin{abstract}
Generating realistic 3D head motion for dyadic interactions is a significant challenge in virtual human synthesis. While recent methods achieve impressive results with speaking heads, they frequently suffer from the `Regression-to-the-Mean' problem in listener motions, collapsing into static faces, and lack the parameter space for complex nonverbal motions. In this paper, we propose GDPO-Listener, a novel framework that achieves highly expressive speaking and listening motion generation. First, we introduce an Auto-Regressive Flow Matching architecture enabling stable supervised learning. Second, to overcome kinematic stillness, we apply the Group reward-Decoupled Policy Optimization (GDPO). By isolating reward normalization across distinct FLAME parameter groups, GDPO explicitly incentivizes high variance expressive generations. Finally, we enable explicit semantic text control for customizable responses. Extensive evaluations across the Seamless Interaction and DualTalk datasets demonstrate superior performance compared to existing baselines on long-term kinematic variance, visual expressivity and semantic controllability.
\end{abstract}

\section{Introduction}
\label{sec:intro}

Conditional speaking/listening head generation is essential for building socially intelligent systems capable of establishing interpersonal trust through nonverbal behaviors \cite{rapport}. While traditional audio-driven models have achieved accurate speaker lip synchronization, they frequently neglect the listener's nuanced nonverbal feedback. Recent dyadic methods attempt to model these coordinated reactions, yet they typically restrict movements to short temporal windows and rely on simplified motion tokenization rather than high-fidelity expressions \cite{tran2024dim, ng2022learning, liu2024customlistener, fan2022faceformer, richard2021meshtalk, xing2023codetalker}. 
Despite recent advancements, generating highly expressive and naturalistic motions during dyadic interactions still remains challenging due to the following difficulties:

\textbf{`Regression-to-the-Mean' Problem in Listener Head Learning}.
Speaking head generation is a highly deterministic, one-to-one mapping problem strongly controlled by the speaker's audio. For instance, if the input audio is the word `ouch', the avatar must open its mouth widely, likely accompanied by a painful expression. Because the physical ground truth is strictly bound to the audio signal, supervised learning paradigms easily and accurately solve this problem.
In contrast, listening head generation is non-deterministic~\cite{zhou2022responsive}. When the actor's own audio is silent, its motion is only weakly correlated with the partner's audio and motion, yielding an inherently one-to-many correspondence~\cite{ng2022learning, wang2025diffusion, ki2026avatar}. For example, if the speaker asks, `Did your paper get accepted?', the listener might exhibit an excited, exaggerated nod (expressing `yes'), or a sad, slow head shake (expressing `no'). Both are entirely valid, natural responses to the exact same conversational context.
Current listener generation methods all fundamentally treat this as a standard supervised learning task, whether they rely on pure regression (DIM, DualTalk~\cite{tran2024dim, peng2025dualtalk}), diffusion (DiffPoseTalk, Infp, CustomListener~\cite{sun2024diffposetalk,zhu2025infp, liu2024customlistener}), or auto-regressive models (ARTalk~\cite{chu2025artalk}, UniLS~\cite{chu2025unils}). They use either MSE loss in continuous spaces or Cross Entropy loss in discrete spaces, but the objective is always to minimize the error against a single ground-truth trajectory. When a network is forced to map a single input to a highly diverse set of valid outputs, it mathematically optimizes for the average of all possible futures. Consequently, the model collapses into predicting safe, low-variance, `mean' motions. This `Regression-to-the-Mean' physically manifests as a stiff, static, and unnatural listening avatar.
To bypass this regression, recent state-of-the-art methods like UniLS \cite{chu2025unils} propose complex two-stage supervised learning pipelines: first learning an audio-free motion prior, followed by an audio-conditioned fine-tuning stage. However, this introduces critical flaws. First, the audio-free training stage is heavily biased toward the dominant speaking motions in the dataset, failing to learn listening priors. Second, the subsequent audio-conditioned fine-tuning stage inevitably suffers from catastrophic forgetting, degrading the expressiveness learned in the first stage.
Instead of relying on multi-stage supervised workarounds, we treat it as an alignment problem. We introduce a post-training paradigm using Group reward-Decoupled Policy Optimization (GDPO) \cite{liu2026gdpo}. After standard supervised training, we post-train the model using sequence-level rewards specifically designed to maximize motion diversity and expressive dynamics, avoiding the average result and restoring natural human expressiveness.

\begin{figure}[t!]
  \centering
  \includegraphics[height=5cm]{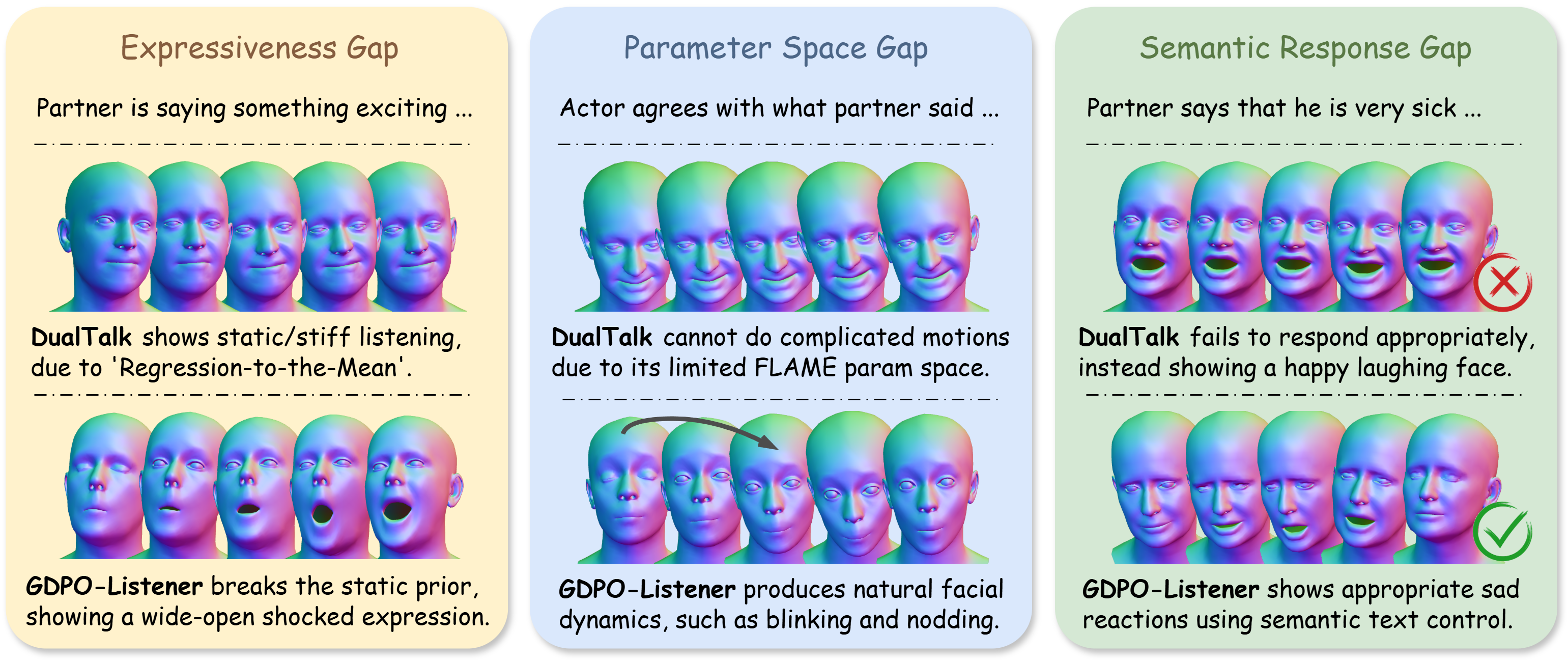}
  \caption{
  \textbf{Necessity of Our Method}.
We resolve three critical gaps in previous baselines. (Left) Older methods suffer from static mean-collapse, but we synthesize high-variance expressive reactions. (Middle) Baselines cannot produce complex motions, whereas we enable natural eye blinks and head nods. (Right) Prior models often have no semantic guidance, but our text control ensures contextually appropriate responses.
  }\vspace{-1em}
  \label{fig:example}
\end{figure}

\textbf{Lack of Expressiveness in Listener Head Generation}.
Current listening head models struggle to synthesize expressive, contextually appropriate responses.
For example, they frequently produce unreal motion artifacts (\textit{e.g.}, unnatural head/jaw movements or high-frequency jitter).
And they also suffer from severe contextual misalignment, where the avatar generates negative expressions (\textit{e.g.}, sadness) in response to positive semantic cues (\textit{e.g.}, hearing `You passed the exam!'). 
This lack of natural expressiveness comes from four critical architectural and methodological weaknesses:
\textit{(a) Data Imbalance}.
Natural dyadic datasets are dominated by neutral, low-energy listening states. By training uniformly across this imbalanced data, previous models statistically overfit to a static, low-amplitude prior, suppressing rare but critical expressive reactions (\textit{e.g.}, enthusiastic nodding or shock).
\textit{(b) Restricted FLAME Parameter Spaces}.
Existing methods typically predict only a limited subset of FLAME parameters (expression, jaw, and neck pose). Omitting eyelid articulation and global head rotation renders them physically incapable of generating essential behaviors like timely blinks, widened eyes, or large-amplitude head nods.
\textit{(c) Limited Explicit Semantic Guidance}.
Audio features alone lack the high-level cognitive context required to determine how a listener should react, leading to inappropriate responses. 
While some recent methods (\textit{e.g.}, UniLS \cite{chu2025unils}) attempt to inject diversity using latent style embeddings, these vectors are highly uninterpretable black boxes that require blind trial-and-error sampling or reference videos during inference.
\textit{(d) Absence of Expressiveness Control}.
Real-world applications require control over the magnitude of a reaction.
Previous generative baselines have a limited ability to modulate expression intensity over time, making it impossible to smoothly transition between a slight smile and an exaggerated laugh. 
In this work, we systematically address these four limitations to achieve highly expressive, intent-driven listener head generations.

To overcome these barriers, we present \textbf{GDPO-Listener}, a novel co-speech listener head generative framework designed to synthesize highly expressive and contextually appropriate listening behaviors.
We formulate the base generation as an Auto-Regressive Flow Matching (AR-Flow) process, which seamlessly adapts for long sequences and multimodal prompting. It is further empowered by Classifier-Free Guidance (CFG)~\cite{ho2022classifier} to support expression intensity control.
Crucially, to overcome the `Regression-to-the-Mean' problem and the inherent static bias of the training data, we propose a GDPO-based post-training scheme.
It directly rewards dynamic motion to force the model out of its static, low-variance prior.
By decoupling these rewards across the different numerical scales of the FLAME parameters, we ensure balanced expressiveness without dominant motions taking over.
Next, to break the physical bottlenecks of previous baselines, we expand the predictive FLAME parameter space by adding eyelid articulation, eye pose, and global head rotation to our variational autoencoder (VAE) \cite{Kingma2013}.
By further conditioning our model on explicit text prompts, we support interpretable, zero-shot semantic control over the listener's cognitive intent.
Extensive evaluations on the Seamless Interaction \cite{agrawal2025seamless} and DualTalk datasets demonstrate that our framework achieves competitive metrics and superior motion quality compared to prior methods.
In summary, our main contributions are:

\textit{(a)} An Auto-Regressive Flow Matching interactive (speaker/listener) head generation framework that accepts multimodal prompts and explicit semantic or intensity controls.

\textit{(b)} A Group reward-Decoupled Policy Optimization post-training reinforcement learning technique with rewards directly optimized for natural, highly expressive listener heads.

\section{Related Works}
\label{sec:related_works}

\textbf{Dyadic Conversational Head Generation}.
Talking head models predict facial motion from strongly correlated audio, but listening head generation is fundamentally different. Listener reactions are weakly correlated with the speaker's audio, making the mapping highly stochastic. Supervised models thus suffer from the `Regression-to-the-Mean' problem, producing static and low-expressive predictions.
Early dyadic approaches used pure regression (DIM, DualTalk \cite{tran2024dim, peng2025dualtalk}) or diffusion (DiTaiListener, Infp, Audio-driven 3D avatars) \cite{siniukov2025ditailistener,zhu2025infp,Ng_2024_CVPR}. Relying on fixed-window forward passes, they require extra temporal smoothing and struggle with real-time, long-sequence generation. Pure Auto-Regressive (AR) models, like UniLS \cite{chu2025unils} or LM-Listener \cite{Ng_2023_ICCV}, solve this by causally predicting frames, natively supporting infinite sequences.
However, concurrent AR methods fail in their representation space. The field uses FLAME to parameterize continuous 3D facial kinematics.  Recent AR models tokenize these movements into discrete codebooks via VQ-VAE~\cite{van2017neural} or BSQ-VAE~\cite{zhao2024image}. Forcing continuous data into discrete spaces introduces severe quantization errors. In facial animation, this manifests as unnatural jitter, abruptly snapping between states.
To maintain continuous human kinematics, we then employ Auto-Regressive Flow Matching, combining causal AR generation with the mathematically high-fidelity synthesis of continuous latent flows.

\textbf{Resolving the `Regression-to-the-Mean' Problem}.
Listener reactions are weakly correlated with the speaker's audio and motion. To minimize supervised loss, models select static and low-expressive faces. Several methods attempt to resolve this one-to-many mapping.
Unlike LLMs, which avoid this by leveraging trillions of data points and categorical cross-entropy, 3D facial datasets lack the volume to naturally overcome this averaging effect.
Some works frame the issue as an exposure bias between teacher-forcing training and auto-regressive inference \cite{zhou2025taming}. They apply techniques like Scheduled Sampling, or in continuous generative spaces, Diffusion-Forcing and Self-Forcing. Others, such as UniLS \cite{chu2025unils}, employ two-stage training: unconditional free-motion followed by conditional dyadic finetuning.
Another direction is to strengthen the weak conditioning to force a specific output. DiTaiListener injects semantic text control, while UniLS and ARTalk utilize explicit style embeddings.
However, these solutions fail to address the core mechanism. They still fundamentally rely on supervised learning from ground truth trajectories for a one-to-many generative task. Instead, we treat listening head generation as an alignment problem. By applying Group reward-Decoupled Policy Optimization (GDPO), we abandon the restrictive trajectory-matching approach. We optimize a specialized reward function, achieving highly expressive and high-fidelity listener reactions without structural degradation.

\textbf{Reinforcement Learning for Listening Head Generation}.
While prior head generation methods rely on supervised trajectory matching~\cite{ng2022learning, sun2024diffposetalk, yu2023talking}, RLHF~\cite{ouyang2022training} and preference-based optimization (\textit{e.g.}, PPO, DPO, GRPO) provide a compelling alternative to directly optimize perceptual or task-driven objectives~\cite{schulman2017proximal, rafailov2023direct, shao2024deepseekmath}. 
Importantly, these post-training paradigms have started to move beyond discrete-token LLMs to continuous generative models, including diffusion for images and videos~\cite{wu2025densedpo, wallace2024diffusion, liu2025videodpo}, and have very recently been explored in talking-head animation as well~\cite{tan2026flowportraitreinforcementlearningaudiodriven}. 
However, RL-style post-training remains largely unexplored in dyadic listening head generation, and directly applying preference optimization to 3D facial kinematics is non-trivial. Heterogeneous parameter subspaces with vastly different scales and dimensionalities can destabilize naive reward aggregation. In dyadic listener head generation, where multiple reactions can be equally valid for the same context, MSE-style trajectory matching can penalize expressive yet appropriate responses, and even continuous adaptations such as Diffusion-DPO~\cite{wallace2024diffusion} may still favor conservative outputs when preference signals are trajectory-agnostic in such multimodal settings.
Recent flow-model post-training (\textit{e.g.}, Flow-GRPO~\cite{liu2025flow}) show that exploration-aware rollouts are critical for stable group-relative policy updates, as deterministic sampling can yield insufficient intra-group diversity to estimate advantages reliably. Yet, none of these works address the unique challenges posed by 3D facial kinematics in dyadic settings, a gap our method is designed to address.


\begin{figure}[t!]
  \centering
  \includegraphics[height=10.5cm]{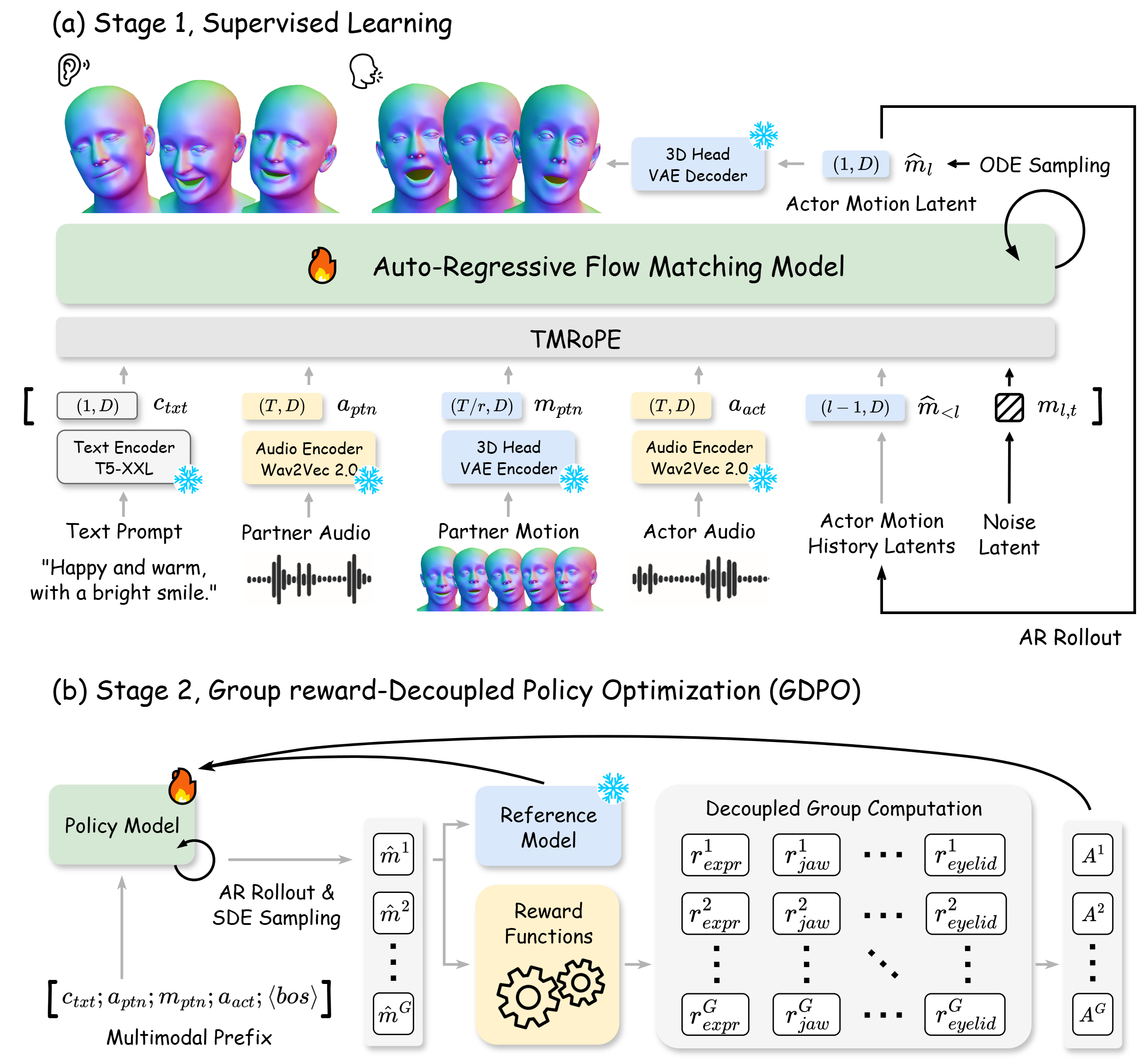}
  \caption{
  \textbf{GDPO-Listener Architecture}.
  Our framework has two training stages. (a) Supervised Learning. Multimodal inputs are encoded as prefix conditions, and an Auto-Regressive Flow Matching model iteratively predicts actor motion latents from noise and history via ODE sampling. (b) Reinforcement Learning. We then post-train the policy model via GDPO. We compute fine-grained, decoupled rewards for distinct FLAME parameters under SDE sampling to explicitly optimize expressiveness.
  }
  \label{fig:main_arch}
\end{figure}

\section{Methods}
\label{sec:methods}

An overview of our \textbf{GDPO-Listener} framework is presented in Fig. \ref{fig:main_arch}. 
We design a two-stage training mechanism to synthesize highly expressive listener head reactions.
The following sections provide further details: Sec. \ref{sec:method_preliminaries} introduces our continuous facial kinematics representations. Sec. \ref{sec:method_arflow} explains our Auto-Regressive Flow Matching base model. Finally, Sec. \ref{sec:method_gdpo} details Group reward-Decoupled Policy Optimization (GDPO), our reinforcement learning stage applied exclusively to listener frames to optimize expressive rewards.

\subsection{Preliminaries}
\label{sec:method_preliminaries}

We parameterize facial kinematics using the FLAME 3D morphable model (3DMM) \cite{li_learning_2017}. Previous dyadic methods restrict this space to low-dimensional representations, with separate expression and pose (jaw and neck) parameters. To capture the full spectrum of nonverbal cues, we expand the motion space by adding eyelid closure, eye pose, and global head rotation. This enables the synthesis of expressive behaviors and realistic interactions, such as eye blinking and intense head nodding.

Let $M \in \mathbb{R}^{L \times d}$ denote a continuous motion sequence. Recent listener generation models \cite{tran2024dim,ng2022learning}, compress $M$ using discrete quantization (\textit{e.g.}, VQ-VAE). However, forcing continuous kinematics into discrete codebooks introduces quantization bounds, physically manifesting as high-frequency temporal jitter. To preserve smooth human motion, we employ a transformer-based VAE to learn a continuous latent space.
For stable optimization, we first normalize the FLAME parameters channel-wise to obtain $M'$.
The VAE encoder $\mathcal{E}_{\text{enc}}$ then maps $M'$ into a temporally compressed distribution $m \in \mathbb{R}^{L/r \times D_v}$, where $r$ is the temporal downsampling rate.

\subsection{Stage 1: Supervised Learning}
\label{sec:method_arflow}
In this stage, we employ a supervised auto-regressive flow matching formulation to establish a high-fidelity base policy for both speaker and listener head generation, ensuring strict adherence to multimodal constraints.
We first extract embeddings for the text prompt, the partner's audio and motion, and the actor's audio. These are projected into dimension $D$ to form the conditioning prefix:
\begin{equation}
\begin{aligned}
c_{\text{txt}} &= (\mathcal{P}_{\text{txt}} \circ \mathcal{T}_{\text{enc}})(\texttt{text\_prompt})\\
[a_{\text{act}}; a_{\text{ptn}}] &= (\mathcal{P}_{\text{audio}} \circ \mathcal{A}_{\text{enc}})([\texttt{actor\_audio}; \texttt{partner\_audio}])\\
m_{\text{ptn}} &= (\mathcal{P}_{\text{motion}} \circ \mathcal{E}_{\text{enc}})(\texttt{partner\_motion})
\end{aligned}
\end{equation}
where $c_{\text{txt}} \in \mathbb{R}^{1 \times D}$, $a_{\text{act}}, a_{\text{ptn}} \in \mathbb{R}^{L \times D}$, and $m_{\text{ptn}} \in \mathbb{R}^{L/r \times D}$. Here, $L$ is the sequence length, $r$ is the temporal downsampling rate, $\mathcal{P}_{*}$ are trainable linear projectors, and the upstream encoders ($\mathcal{T}_{\text{enc}}, \mathcal{A}_{\text{enc}}, \mathcal{E}_{\text{enc}}$) remain frozen.

\textbf{Time-aligned Multimodal RoPE (TMRoPE)}.
Precise audio-visual synchronization is critical for realistic speaking and listening head generation.
And it is even more challenging due to various temporal resolutions of multimodal inputs (\textit{e.g.}, high-frequency audio versus downsampled motion latents).
To resolve this temporal ambiguity, we use TMRoPE, inspired by multimodal alignment in Qwen2.5-Omni \cite{xu2025qwen25omnitechnicalreport}. We map all input streams into a shared temporal coordinate system by assigning an explicit timestamp index $t$ to each token:
\begin{equation}
\left\{
\begin{aligned}
&t^{\text{text}} = 0 \\
&t^{\text{audio}}_i = i, \quad &&i \in \{0, 1, \dots, L-1\} \\
&t^{\text{motion}}_j = j \cdot r + \lfloor r/2 \rfloor, \quad &&j \in \{0, 1, \dots, \lfloor L/r\rfloor- 1\}
\end{aligned}
\right.
\end{equation}
where $L$ is the audio sequence length and $r$ is the motion downsampling rate. The offset $\lfloor r/2 \rfloor$ mathematically anchors each low-frequency motion token to the exact center of its corresponding high-frequency audio window. Within the transformer blocks, we apply Rotary Position Embeddings (RoPE)~\cite{su2024roformer} with aforementioned modality-specific indices to self-attention input, equipping the model with strict, physically grounded cross-modal alignment.

\textbf{Auto-Regressive Flow Matching (AR-Flow)}.
We model the generation of the $l$-th motion latent token $m_l$ via continuous flow matching. Let $t \in [0, 1]$ denote the flow timestep, where $m_{l,0} \sim \mathcal{N}(0, I)$ is Gaussian noise and $m_{l,1} = m_l$ is the $l$-th ground truth latent.
The intermediate state is defined via linear interpolation $m_{l,t} = t\cdot m_l + (1-t)m_{l,0}$. The AR-Flow network $\mathcal{F}$ predicts the optimal velocity vector $v_\theta$ pointing toward the target:
\begin{equation}
v_\theta(m_{l,t}) = \mathcal{F}([c_{\text{txt}}; a_{\text{ptn}}; m_{\text{ptn}}; a_{\text{act}}; \hat{m}_{<l}; m_{l,t}], t)
\end{equation}
The model is optimized using the Rectified Flow objective, minimizing:
\begin{equation}
\mathcal{L}_{\text{FM}} = \mathbb{E}_{t , m_l, m_{l,0}}\left[ \| v_\theta(m_{l,t}) - (m_l - m_{l,0}) \|^2 \right]
\end{equation}
During inference time, we sample the latent token by solving the Ordinary Differential Equation (ODE) using Euler sampler from $t=0$ to $t=1$.
\begin{equation}
\hat{m}_{l,t+\Delta t} = \hat{m}_{l,t} + \Delta t \cdot v_\theta(\hat{m}_{l,t})
\end{equation}
Once the $l$-th motion latent $\hat{m}_{l}$ is fully estimated, it will be appended to the actor's motion history, such that $\hat{m}_{<(l+1)}=[\hat{m}_{<l};\hat{m}_{l}]$. 
This updated history then conditions the AR-Flow network $\mathcal{F}$ for the subsequent latent prediction, enabling auto-regressive generation.
For inference-time expressiveness control, we use Classifier-Free Guidance (CFG). We extrapolate the inference velocity with an unconditional prior $v_\emptyset(\hat{m}_{l,t})$ and the conditional prior $v_{\text{cond}}(\hat{m}_{l,t})$:
\begin{equation}
\hat{v}_\theta(\hat{m}_{l,t}) = v_\emptyset(\hat{m}_{l,t}) + \omega_l \cdot \left( v_{\text{cond}}(\hat{m}_{l,t}) - v_\emptyset(\hat{m}_{l,t}) \right)
\end{equation}
where $\omega_l \in \{\omega_{\text{speak}}, \omega_{\text{listen}}\}$ based on whether the actor is speaking or listening at frame $l$. This decoupling enables independent tuning of active lip-sync fidelity and reactive listener variance without retraining.

\subsection{Stage 2: Reinforcement Learning}
\label{sec:method_gdpo}

Supervised baselines inevitably suffer from the aforementioned `Regression-to-the-Mean' problem, averaging one-to-many listener reactions into static, expressionless faces. To break this deterministic collapse and drive more expressive listener-head generation, we use Group reward-Decoupled Policy Optimization \cite{liu2026gdpo} as a post-training stage for our autoregressive flow-matching framework.

\textbf{From ODE to SDE}.
Group-based RL requires stochastic exploration, but our baseline, AR-Flow, employs deterministic ODE sampling: $dm_{l,t} = v_\theta(m_{l,t})dt$. 
To enable policy optimization, we inject necessary randomness by converting the generation process into a continuous Stochastic Differential Equations (SDE): $dm_{l,t} = v_\theta(m_{l,t})dt + \sigma dw_t$. 
Applying Euler-Maruyama discretization yields our practical update rule:
\begin{equation}
\hat{m}_{l,t+\Delta t} = \hat{m}_{l,t} + \Delta t\cdot v_{\theta}(\hat{m}_{l,t}) + \sigma \sqrt{\Delta t} \cdot \epsilon
\end{equation}
where $\epsilon \sim \mathcal{N}(0,I)$ provides exploratory variance and $\sigma$ is the hyper-parameter constant controlling the noise scale. Unlike standard Flow-GRPO \cite{liu2025flow}, we strategically drop the score-matching drift correction to reduce computational overhead during online RL rollouts. Furthermore, we use significantly fewer denoising steps during training for efficiency. At test time, we revert to the fast, deterministic ODE sampler, achieving high-fidelity generation without test-time computational overhead.

\textbf{Group reward-Decoupled Policy Optimization (GDPO)}.
Applying group-based online RL methods to listener head generation presents unique challenges due to the multi-dimensional nature of the motion space. We optimize over $N$-dimensional distinct FLAME parameters, indexed by $i \in \{\texttt{expr}, \texttt{jaw}, \texttt{neck}, \\\texttt{eyelid}, \texttt{eyepose}, \texttt{rot}\}$. 
To maintain the distinctions among diverse reward combinations and accurately capture their relative scales, we adapt the GDPO.
We first execute the AR rollout and SDE sampling to generate a full sequence. This yields a trajectory set $\{\hat{m}_l^j\}_{l=1}^{L/r}$, and $L/r$ denotes the latent length. 
For each candidate trajectory, we compute sequence-level rewards $R_i^j$ by comparing the generated sequence against the ground truth $\{m_l^j\}_{l=1}^{L/r}$.
To prevent large-magnitude rewards from overshadowing penalties, we compute normalized advantages independently for each sample $j\in\{1,2,\dotsc,G\}$ within the group:
\begin{equation}
A_i^j = \frac{R_i^j - \text{mean}(\{R_i^k\}_{k=1}^G)}{\text{std}(\{R_i^k\}_{k=1}^G)},\quad R_i^j = R_i^j(\{\hat{m}_l^j\}_{l=1}^{L/r}, \{m_l^j\}_{l=1}^{L/r})
\end{equation}
The final scalar advantage for the $j$-th rollout is then obtained via a normalized weighted sum across the $N$ FLAME parameter spaces:
\begin{equation}
A^j = \frac{\sum_{i=1}^N \lambda_i A_i^j}{\sum_{i=1}^N \lambda_i + \epsilon_0}
\end{equation}
where $\lambda_i$ provides explicit control over the relative importance of specific FLAME parameters, and $\epsilon_0$ ensures numerical stability during computation.
Ultimately, the policy $\pi_\theta$ is updated by maximizing the task-specific surrogate objective:
\begin{equation}
\mathbb{E} \left[ \frac{1}{G} \sum_{j=1}^G \frac{1}{T} \sum_{t=1}^T \bigg( \min \Big( \rho^j_t A^j, \text{clip}(\rho^j_t, 1-\epsilon, 1+\epsilon) A^j \Big) - \beta D_{\text{KL}}(\pi_\theta || \pi_{\text{ref}}) \bigg) \right]
\end{equation}
where $\rho^j_t = p_\theta/p_{\theta_{\text{old}}}$ is the probability ratio between the current and old policies, and $T$ is the number of SDE denoising steps used during rollouts.
To prevent memory exhaustion from the massive AR-SDE computational graph, we decouple generation from optimization. We first sample and cache full trajectories in a no-gradient mode, and subsequently update the policy by replaying these cached states via chunked backpropagation.

\textbf{Reward Function}.
To explicitly break the `Regression-to-the-Mean' collapse and drive highly expressive listener generation, we design a multi-component reward function. 
We decode $\hat{m}$ with the VAE decoder to obtain $\hat{M}$ in FLAME space, then compute per-group rewards on $\hat{M}$.
For a given FLAME parameter feature $i$, the reward $R_i$ synthesizes four distinct objectives: a positional variance penalty $\sum_{l=1}^L|\sigma(\hat{M}_l[i])-\sigma(M_l[i])|$ to match structural motion amplitude, a velocity variance penalty $\sum_{l=1}^L|\sigma(\nabla \hat{M}_l[i]) - \sigma(\nabla M_l[i])|$ to capture high-frequency human dynamics, a mean-drift penalty $|\mu(\hat{M}[i]) - \mu(M[i])|$ to prevent global drift away from the segment’s typical pose, and a standard MSE loss $\sum_{l=1}^L||\hat{M}_l[i]-M_l[i]||^2$ to ensure semantic grounding. These individual penalties are aggregated via a weighted sum to compute the final reward for each FLAME feature term.
Importantly, the reward primarily emphasizes dynamics-aware statistics (variance and velocity variance) rather than per-frame alignment. Combined with group-decoupled normalization, it encourages diverse, high-variance rollouts while using the mean and MSE anchors only to stabilize realism and avoid degenerate drift.

\section{Experiments}
\label{sec:exps}

\subsection{Experiment setting} 
\textbf{Implementation Details}.
Our VAE ($d=67$) compresses $L=200$ frames of $N=6$ FLAME parameter groups ($\texttt{expr}, \texttt{jaw}, \texttt{neck}, \texttt{eyelid}, \texttt{eyepose}, \texttt{rot}$). With downsampling $r=8$, the latent length is 25. AR-Flow is a continuous Transformer (6 layers, 16 heads, $D=512$) conditioned on frozen T5-XXL and Wav2Vec2.0-Large. We train on 8 L40s GPUs using AdamW. 
Stage 1 trains the base AR-Flow (lr=$10^{-4}$, batch size 32, 50 epochs). Stage 2 post-trains via GDPO (lr=$5\times 10^{-6}$, batch size 4, 2 epochs). For GDPO, we set group size $G=4$, SDE variance $\sigma=0.5$, KL penalty $\beta=0.01$, replay chunk step to 2. We synchronize the behavior policy every 8 iterations for ratio computation, while keeping the reference policy fixed. To accelerate RL simulation, generation trajectories use 4 flow denoising steps during training, while 10 steps are used for inference.

\begin{table}[t]
\centering
\caption{
Quantitative Analysis on Seamless Interaction Dataset. $*$ means adapting the ARTalk for speak-listen generation by adding additional partner audio input.
}\vspace{-0.5em}
\begin{tabular}{l|ccccc|ccc}
\hline
                          & \multicolumn{5}{c|}{Speaking}                                                                                                                                                                             & \multicolumn{3}{c}{Listening}                                                                                            \\
\multirow{-2}{*}{Methods} & LVE $\downarrow$                                  & MHD $\downarrow$                                  & FDD  $\downarrow$                                  & PDD  $\downarrow$                                 & JDD  $\downarrow$                                 & FDD  $\downarrow$                                  & PDD  $\downarrow$                                & JDD  $\downarrow$                                 \\ \hline
DiffposeTalk~\cite{sun2024diffposetalk}              & 9.48                                  & 2.96                                  & 32.66                                  & 7.89                                  & 1.40                                   & -                                      & -                                    & -                                     \\
L2L~\cite{ng2022learning}                       & 3.10                                      &  1.25                                     &  53.03                                      & 3.52                                      &  3.32                                     &  25.95                                      &  2.18                                    &  1.14                                      \\
ARTalk~\cite{chu2025artalk}                    & 7.46                                  & 2.12                                  & 31.64                                  & 7.66                                  & 1.19                                  & -                                      & -                                    & -                                     \\
ARTalk*~\cite{chu2025artalk}                   & 6.79                                  & 2.02                                  & 27.41                                  & 8.55                                  & \cellcolor[HTML]{EFEFEF}\textbf{0.81} & 30.62                                  & 9.52                                 & 1.53                                  \\
DualTalk~\cite{peng2025dualtalk}                  & 6.53                                  & 1.95                                  & 37.46                                  & 9.70                                   & 1.02                                  & 43.58                                  & 10.71                                & 2.02                                  \\ \hline
\rule[-5pt]{0pt}{14pt}\textbf{GDPO-Listener}             & \cellcolor[HTML]{EFEFEF}\textbf{2.95} & \cellcolor[HTML]{EFEFEF}\textbf{1.06} & \cellcolor[HTML]{EFEFEF}\textbf{22.45} & \cellcolor[HTML]{EFEFEF}\textbf{2.71} & 1.26                                  & \cellcolor[HTML]{EFEFEF}\textbf{18.85} & \cellcolor[HTML]{EFEFEF}\textbf{1.90} & \cellcolor[HTML]{EFEFEF}\textbf{0.87} \\ \hline
\end{tabular}
\label{tab:full_seamless}

\bigskip 

\centering
\caption{
Quantitative Analysis on Seamless Interaction Expressive Subset.
}\vspace{-0.5em}
\begin{tabular}{l|ccccc|ccc}
\hline
                          & \multicolumn{5}{c|}{Speaking}                                                                                                                                                                             & \multicolumn{3}{c}{Listening}                                                                                            \\
\multirow{-2}{*}{Methods} & LVE $\downarrow$                                  & MHD $\downarrow$                                  & FDD  $\downarrow$                                  & PDD  $\downarrow$                                 & JDD  $\downarrow$                                 & FDD  $\downarrow$                                  & PDD  $\downarrow$                                & JDD  $\downarrow$                                 \\ \hline
L2L~\cite{ng2022learning}                       &   3.40                                    &  1.52                                    &  55.10                                      &   5.15                                    & 3.38                                     &  44.77                                      &  3.77                                    & 2.39                                      \\
DualTalk~\cite{peng2025dualtalk}                       &     \cellcolor[HTML]{EFEFEF}\textbf{2.86}                                  &  1.34                                     & 64.84                                       &   5.04                                    &    3.70                                   &  55.94                                      &  5.08                                    &  2.79                                     \\ \hline
\rule[-5pt]{0pt}{14pt}\textbf{GDPO-Listener}             & 3.08	& \cellcolor[HTML]{EFEFEF}\textbf{1.25}	& \cellcolor[HTML]{EFEFEF}\textbf{30.87}	& \cellcolor[HTML]{EFEFEF}\textbf{4.44}	& \cellcolor[HTML]{EFEFEF}\textbf{1.67}	& \cellcolor[HTML]{EFEFEF}\textbf{22.56}	& \cellcolor[HTML]{EFEFEF}\textbf{3.33}	& \cellcolor[HTML]{EFEFEF}\textbf{1.48} \\ \hline
\end{tabular}
\label{tab:expr_seamless}
\end{table}

\textbf{Dataset}. 
We utilize the Seamless Interactions dataset of dyadic conversational videos. For data processing, we apply MediaPipe for face detection, extract per-frame FLAME parameters using SHeaP~\cite{schoneveld2025sheap}, and apply Savitzky-Golay filtering~\cite{savitzky1964smoothing} for temporal smoothing. We train on a 718h subset and evaluate on the full 31.5h test split, without introducing any external training data (unlike UniLS). To rigorously evaluate dynamic reactions, we curate an expressive test subset by filtering the test set based on the variance of ground-truth FLAME parameters. Additionally, we evaluate our framework on the DualTalk dataset.


\textbf{Evaluation Metrics}. 
We compare baselines (DiffposeTalk~\cite{sun2024diffposetalk}, L2L~\cite{ng2022learning}, ARTalk~\cite{chu2025artalk}, DualTalk~\cite{peng2025dualtalk}) against our GDPO-Listener.
Our evaluation explicitly distinguishes between speaking accuracy and listening naturalness. For speaking, we assess lip synchronization via Lip Vertex Error (LVE) and overall geometric fidelity using Mean Head Distance (MHD). To quantify temporal consistency, we report Upper-face Dynamic Deviation (FDD), alongside Pose Dynamic Deviation for head (PDD) and jaw (JDD) kinematics. For listening, where direct vertex errors unfairly penalize diverse but plausible reactions, we focus entirely on distribution matching. We compute FDD, PDD, and JDD on listening segments to evaluate how accurately the generated reactive motions align with ground-truth. Details of metrics can be found in the supplementary materials.

\subsection{Quantitative results}

\textbf{Seamless Interaction Dataset}.
We evaluate our framework on the Seamless Interaction dataset. GDPO-Listener significantly outperforms all baselines on the full test set in Tab. \ref{tab:full_seamless}, achieving the lowest vertex errors (LVE 2.95, MHD 1.06) and superior dynamic consistency. Our Listen FDD (18.85) drastically improves upon the auto-regressive DualTalk baseline (43.58). To rigorously evaluate the `Regression-to-the-Mean' problem, we further test on a curated expressive subset featuring high-variance ground-truth motions in Tab. \ref{tab:expr_seamless}. Here, deterministic baselines like DualTalk secure a marginally better LVE (2.86 vs 3.08) by safely collapsing to a static mean. However, they catastrophically fail at distribution matching; our model vastly outperforms DualTalk in expressivity, improving Listen FDD from 55.94 to 22.56 and Listen PDD from 5.08 to 3.33. This quantitative evidence shows that our formulation preserves vivid, lifelike reactions in highly dynamic scenarios.

\textbf{DualTalk Dataset}.
We verify generalization on the DualTalk dataset in Tab. \ref{tab:dualtalk_eval}. GDPO-Listener substantially outperforms the baseline across all dynamic metrics. On the full listening set, our method reduces FDD from 46.36 to 26.95. Crucially, on the high-variance `Expressive Listen' subset, the baseline's mean collapse is highly evident (FDD 53.04), whereas our framework sustains robust distribution matching (FDD 33.99, PDD 3.64). This demonstrates our expressivity improvements successfully generalize to novel data domains.

\begin{table}[t!]
\setlength\tabcolsep{4pt}
\caption{
Quantitative Analysis on DualTalk Dataset \& Expressive Subset.
}\vspace{-0.5em}
\centering
\begin{tabular}{l|ccc|ccc}
\hline
                          & \multicolumn{3}{c|}{All Listening}                                                                                        & \multicolumn{3}{c}{Expressive Listening}                                                                                  \\
\multirow{-2}{*}{Methods} & FDD  $\downarrow$                                  & PDD  $\downarrow$                                 & JDD    $\downarrow$                               & FDD               $\downarrow$                     & PDD      $\downarrow$                             & JDD       $\downarrow$                            \\ \hline
L2L~\cite{ng2022learning}                       &  77.04                                      &   6.24                                    & 4.26                                      &  117.57                                      &   8.57                                    &      6.65                                \\
DualTalk~\cite{peng2025dualtalk}                  & 46.36                                  & 5.07                                  & 2.83                                  & 53.04                                  & 5.01                                  & 3.32                                  \\ \hline
\rule[-5pt]{0pt}{14pt}\textbf{GDPO-Listener}             & \cellcolor[HTML]{EFEFEF}\textbf{26.95} & \cellcolor[HTML]{EFEFEF}\textbf{3.86} & \cellcolor[HTML]{EFEFEF}\textbf{1.62} & \cellcolor[HTML]{EFEFEF}\textbf{33.99} & \cellcolor[HTML]{EFEFEF}\textbf{3.64} & \cellcolor[HTML]{EFEFEF}\textbf{1.94} \\ \hline
\end{tabular}
\vspace{7pt}
\label{tab:dualtalk_eval}
\end{table}



\textbf{Efficiency Analysis}.
Excluding frozen multimodal encoders, our trainable framework is exceptionally lightweight at just 19.3M parameters. This is a fraction of the size of prior baselines like ARTalk (489.5M) and DualTalk (647.3M). Furthermore, by employing our sliding-window inference strategy, our model seamlessly achieves real-time generation.

\textbf{User Studies}. We conducted a comprehensive user study to assess perceptual quality. Evaluators consistently rated GDPO-Listener as superior or highly competitive against baselines in various dimensions. Detailed results can be found in the supplementary materials.
\subsection{Qualitative results}

\begin{figure}[t!]
  \centering
  \includegraphics[height=6.5cm]{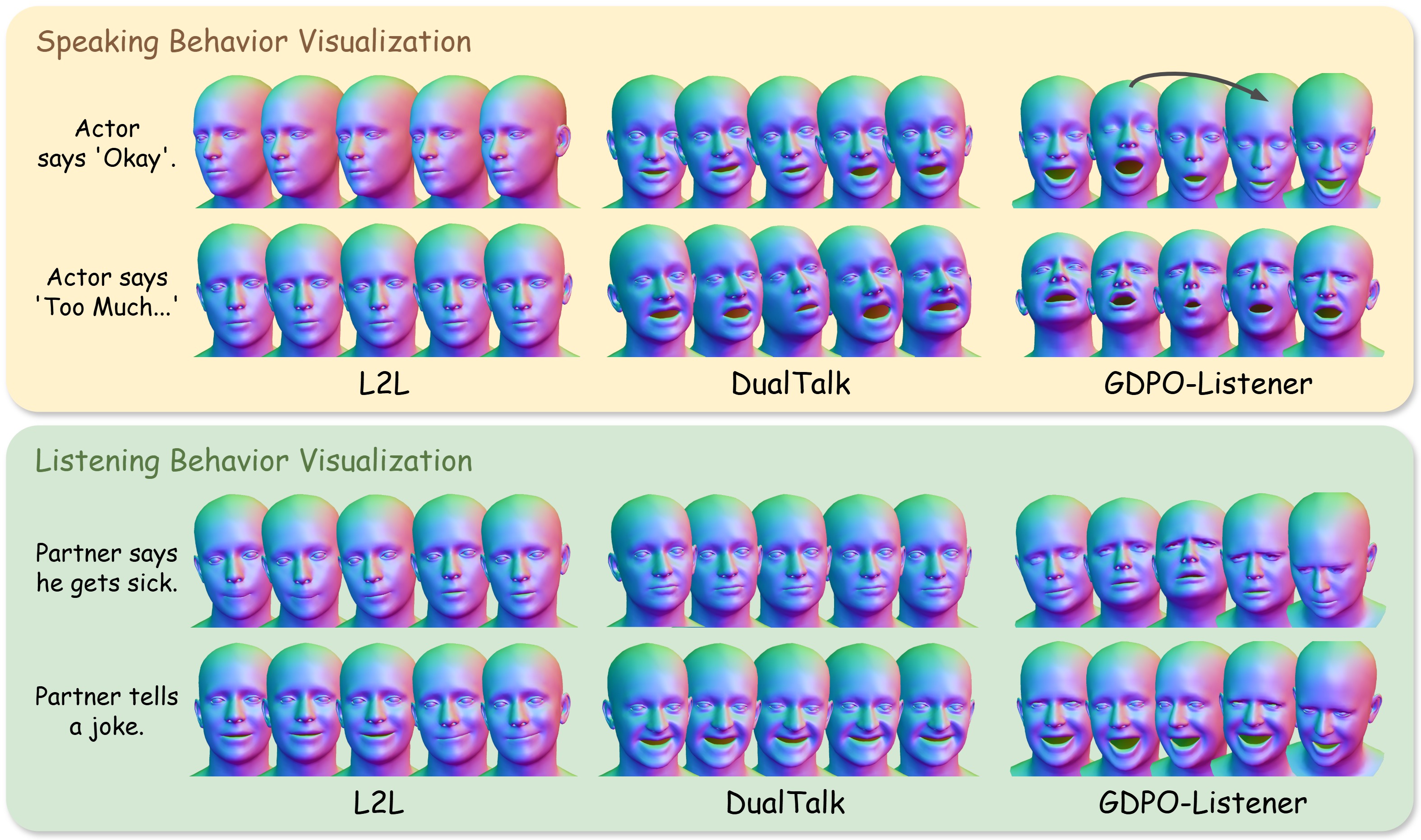}
  \caption{
\textbf{Qualitative Comparisons}. Other methods have low-expressive speaking and static listening, our method shows better lip sync and highly expressive reactions.
  }
  \label{fig:main_compare}

\captionof{table}{
Ablation studies on different components in our framework.
}\vspace{-0.5em}
\medskip

\begin{tabular}{l|ccccc|ccc}
\hline
                          & \multicolumn{5}{c|}{Speaking}                                                                                                                                                                             & \multicolumn{3}{c}{Listening}                                                                                            \\
\multirow{-2}{*}{Methods} & LVE $\downarrow$                                  & MHD $\downarrow$                                  & FDD  $\downarrow$                                  & PDD  $\downarrow$                                 & JDD  $\downarrow$                                 & FDD  $\downarrow$                                  & PDD  $\downarrow$                                & JDD  $\downarrow$                                 \\ \hline
\rule[-5pt]{0pt}{14pt}\textbf{GDPO-Listener}             & 2.95 & 1.06 & \cellcolor[HTML]{EFEFEF}\textbf{22.45} & \cellcolor[HTML]{EFEFEF}\textbf{2.71} & \cellcolor[HTML]{EFEFEF}\textbf{1.26}                                  & \cellcolor[HTML]{EFEFEF}\textbf{18.85} & \cellcolor[HTML]{EFEFEF}\textbf{1.90} & \cellcolor[HTML]{EFEFEF}\textbf{0.87} \\ \hline
\textit{without text prompts} & 2.73 & 1.04 & 33.02 & 2.97 & 1.44 & 22.89 & 2.11 & 0.87 \\ 
\textit{without GDPO} & 3.07 & 1.12 & 39.54 & 3.50 & 1.81 & 28.03 & 2.15 & 1.09 \\ 
\textit{with fewer denoise steps} & \cellcolor[HTML]{EFEFEF}\textbf{2.61} & \cellcolor[HTML]{EFEFEF}\textbf{0.98} & 42.12 & 3.94 & 2.11 & 29.39 & 2.19 & 1.17 \\ 
\hline
\end{tabular}
\label{tab:ablation}
\end{figure}

We visually compare GDPO-Listener against baselines in Fig. \ref{fig:main_compare}. During speaking, L2L and DualTalk exhibit static and low-expressive dynamics, while our method achieves superior lip synchronization and high-variance movements. This gap even widens during listening. When context shifts (\textit{e.g.}, a partner getting sick or joking), baselines suffer from regression to the mean, collapsing into static poses. Conversely, GDPO-Listener overcomes this deterministic prior to synthesize highly expressive, context-aware reactions like empathetic frowning or laughter, ensuring superior visual fidelity and semantic alignment. Additional visualizations are also available in the supplementary materials.

\begin{figure}[t!]
  \centering
  \includegraphics[height=9.5cm]{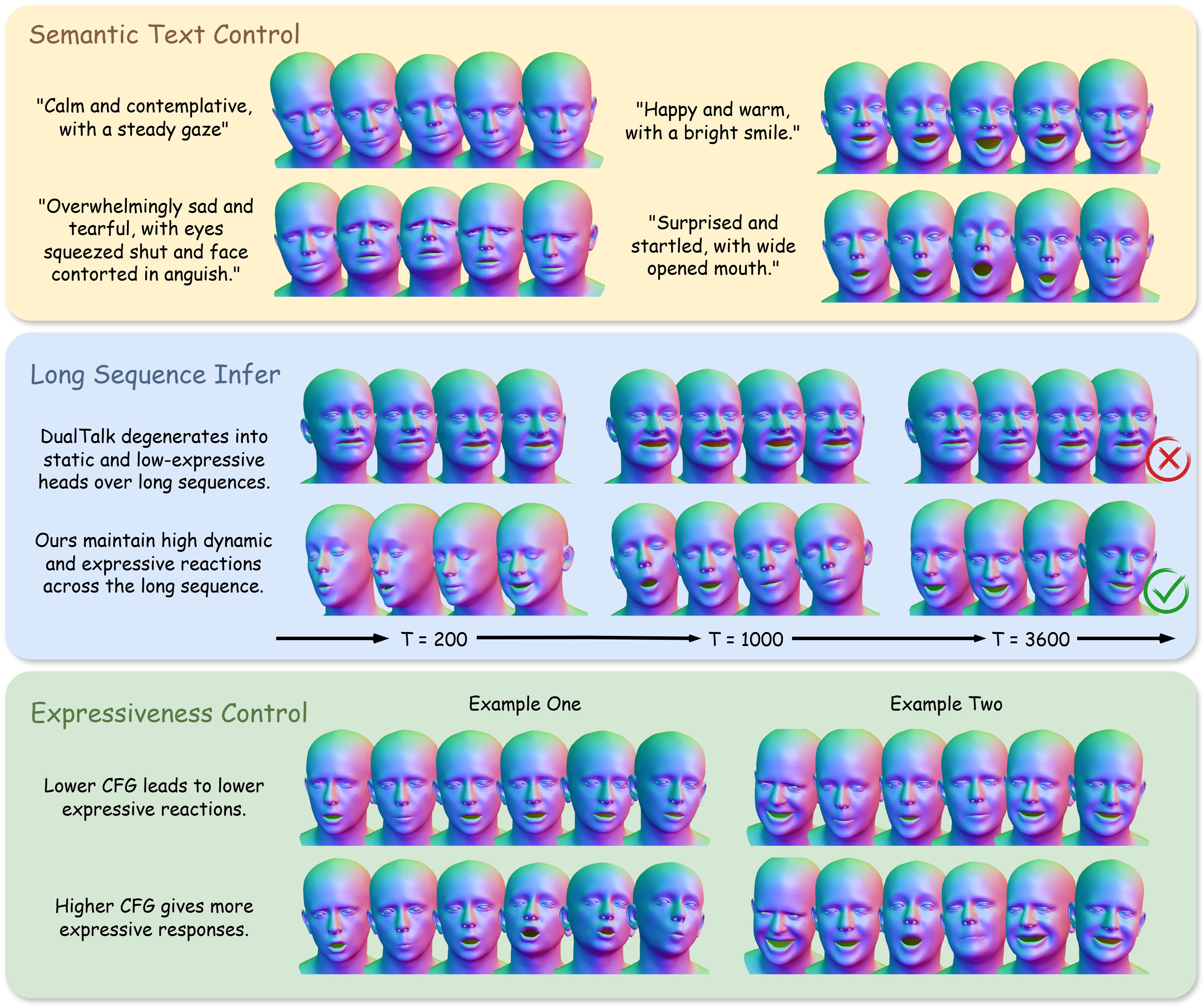}
  \vspace{-0.5em}
  \caption{
  \textbf{Advanced Generation Capabilities}.
(Top) Semantic text explicitly controls emotional states. (Middle) We sustain dynamic reactions during long sequences, avoiding baseline static decay. (Bottom) CFG scaling seamlessly modulates expressiveness intensity without retraining.
  }
  \label{fig:more_visual}
  \vspace{-1em}
\end{figure}

\textbf{Semantic Text Control}. Purely audio-driven baselines often produce inappropriate reactions due to contextual ambiguity (\textit{e.g.}, smiling when hearing `I am sick'). We resolve this using multimodal prefix conditioning for explicit semantic text control. As shown in Fig. \ref{fig:more_visual},  GDPO-Listener seamlessly aligns the listener's kinematics with user-provided text prompts. By supplying descriptions from `happy' to `sad', our model cleanly overrides default acoustic mappings to guarantee contextually appropriate feedback, essential for realistic avatars.

\textbf{Long Sequence Inference}. We employ a sliding-window autoregressive strategy, using the latest $k=1$ latents as a causal prefix for the next window ($W=25$). While methods like DualTalk infer over extended sequences, their regression to the mean amplifies over time, rapidly collapsing into static faces due to purely frame-level objectives. Conversely, GDPO-Listener averts this temporal degradation. By explicitly penalizing variance loss, our model maintains highly dynamic and expressive reactions even at $T=3600$ frames.


\textbf{Expressiveness Control}. We introduce explicit expressiveness control via Classifier-Free Guidance (CFG) scaling. As shown in Fig. \ref{fig:more_visual}, a default CFG of 1.0 matches ground-truth dynamics for evaluation. Without retraining, increasing the CFG scale (\textit{e.g.}, to 4.0) instantly generates significantly more vivid speaking and listening reactions. While previous methods suffer from flat responses or require new data to alter intensity, our approach offers continuous, zero-shot modulation. This acts as a tunable intensity slider, dynamically scaling avatar expressivity for diverse contexts.

\subsection{Ablation Studies}
We ablate our framework's key components in Tab. \ref{tab:ablation}. Crucially, removing GDPO fine-tuning (\textit{without GDPO}) severely degrades all dynamic metrics. 
This shows that supervised flow models suffer from deterministic collapse, underscoring the need for RL optimization to achieve lifelike kinematics. Omitting the text prefix (\textit {without text prompts}) similarly hurts expressiveness, as the model loses semantic guidance to resolve acoustic ambiguities. Finally, reducing flow sampling steps from 10 to 4 (\textit{with fewer denoise steps}) severely penalizes expressivity. This happens because few-step sampling defaults to a safe, mean-collapsed trajectory, which justifies our configuration for balancing rich variance with efficiency.

\section{Conclusions}
\label{sec:conclusion}
Existing listener motion generation models are unable to generate diverse, expressive behaviors, due to human reaction diversity and the ‘Regression-to-the-Mean’ problem. In this paper, we presented a method for alleviating this problem by post-training a generative model to enable lifelike, expressive listener responses. 
Our model can generate both speaking and listening behavior conditioned on conversational partners' motion and speech.
Extensive experiments on DualTalk and Seamless Interaction databases demonstrate the superiority of the proposed approach in terms of expression diversity and motion quality.

%
%
\bibliographystyle{splncs04}
\bibliography{main}

@String(CVPR  = {IEEE Conf. Comput. Vis. Pattern Recog.})

@String(ICCV  = {Int. Conf. Comput. Vis.})

@String(TOG   = {ACM Trans. Graph.})

@String(CVPR  = {CVPR})

@String(ICCV  = {ICCV})

@String(TOG   = {ACM TOG})

@inproceedings{rapport,
	address = {Berlin, Heidelberg},
	title = {Virtual {Rapport}},
	isbn = {978-3-540-37594-4},
	booktitle = {Intelligent {Virtual} {Agents}},
	publisher = {Springer Berlin Heidelberg},
	author = {Gratch, Jonathan and Okhmatovskaia, Anna and Lamothe, Francois and Marsella, Stacy and Morales, Mathieu and van der Werf, R J and Morency, Louis-Philippe},
	editor = {Gratch, Jonathan and Young, Michael and Aylett, Ruth and Ballin, Daniel and Olivier, Patrick},
	year = {2006},
	pages = {14--27},
}

@inproceedings{peng2025dualtalk,
    title={DualTalk: Dual-Speaker Interaction for 3D Talking Head Conversations},
    author={Ziqiao Peng and Yanbo Fan and Haoyu Wu and Xuan Wang and Hongyan Liu and Jun He and Zhaoxin Fan},
    booktitle={Proceedings of the IEEE/CVF Conference on Computer Vision and Pattern Recognition},
    year={2025},
}

@article{sun2024diffposetalk,
  title={DiffPoseTalk: Speech-Driven Stylistic 3D Facial Animation and Head Pose Generation via Diffusion Models},
  author={Sun, Zhiyao and Lv, Tian and Ye, Sheng and Lin, Matthieu and Sheng, Jenny and Wen, Yu-Hui and Yu, Minjing and Liu, Yong-Jin},
  journal={ACM Transactions on Graphics (TOG)},
  doi={10.1145/3658221},
  volume={43},
  number={4},
  articleno={46},
  numpages={9},
  year={2024},
  publisher={ACM New York, NY, USA}
}

@inproceedings{tran2024dim,
  title={Dim: Dyadic interaction modeling for social behavior generation},
  author={Tran, Minh and Chang, Di and Siniukov, Maksim and Soleymani, Mohammad},
  booktitle={European Conference on Computer Vision},
  pages={484--503},
  year={2024},
  organization={Springer}
}

@inproceedings{chu2025artalk,
  title={Artalk: Speech-driven 3d head animation via autoregressive model},
  author={Chu, Xuangeng and Goswami, Nabarun and Cui, Ziteng and Wang, Hanqin and Harada, Tatsuya},
  booktitle={Proceedings of the SIGGRAPH Asia 2025 Conference Papers},
  pages={1--9},
  year={2025}
}

@article{chu2025unils,
  title={UniLS: End-to-End Audio-Driven Avatars for Unified Listening and Speaking},
  author={Chu, Xuangeng and Liu, Ruicong and Huang, Yifei and Liu, Yun and Peng, Yichen and Zheng, Bo},
  journal={arXiv preprint arXiv:2512.09327},
  year={2025}
}

@article{ho2022classifier,
  title={Classifier-free diffusion guidance},
  author={Ho, Jonathan and Salimans, Tim},
  journal={arXiv preprint arXiv:2207.12598},
  year={2022}
}

@inproceedings{siniukov2025ditailistener,
  title={Ditailistener: Controllable high fidelity listener video generation with diffusion},
  author={Siniukov, Maksim and Chang, Di and Tran, Minh and Gong, Hongkun and Chaubey, Ashutosh and Soleymani, Mohammad},
  booktitle={Proceedings of the IEEE/CVF International Conference on Computer Vision},
  pages={11991--12001},
  year={2025}
}

@inproceedings{ng2022learning,
  title={Learning to listen: Modeling non-deterministic dyadic facial motion},
  author={Ng, Evonne and Joo, Hanbyul and Hu, Liwen and Li, Hao and Darrell, Trevor and Kanazawa, Angjoo and Ginosar, Shiry},
  booktitle={Proceedings of the IEEE/CVF conference on computer vision and pattern recognition},
  pages={20395--20405},
  year={2022}
}

@inproceedings{wang2025diffusion,
  title={Diffusion-based realistic listening head generation via hybrid motion modeling},
  author={Wang, Yinuo and Fan, Yanbo and Wang, Xuan and Yu, Guo and Wang, Fei},
  booktitle={Proceedings of the Computer Vision and Pattern Recognition Conference},
  pages={15885--15895},
  year={2025}
}

@article{ki2026avatar,
  title={Avatar Forcing: Real-Time Interactive Head Avatar Generation for Natural Conversation},
  author={Ki, Taekyung and Jang, Sangwon and Jo, Jaehyeong and Yoon, Jaehong and Hwang, Sung Ju},
  journal={arXiv preprint arXiv:2601.00664},
  year={2026}
}

@article{schulman2017proximal,
  title={Proximal policy optimization algorithms},
  author={Schulman, John and Wolski, Filip and Dhariwal, Prafulla and Radford, Alec and Klimov, Oleg},
  journal={arXiv preprint arXiv:1707.06347},
  year={2017}
}

@article{shao2024deepseekmath,
  title={Deepseekmath: Pushing the limits of mathematical reasoning in open language models},
  author={Shao, Zhihong and Wang, Peiyi and Zhu, Qihao and Xu, Runxin and Song, Junxiao and Bi, Xiao and Zhang, Haowei and Zhang, Mingchuan and Li, YK and Wu, Yang and others},
  journal={arXiv preprint arXiv:2402.03300},
  year={2024}
}

@inproceedings{wallace2024diffusion,
  title={Diffusion model alignment using direct preference optimization},
  author={Wallace, Bram and Dang, Meihua and Rafailov, Rafael and Zhou, Linqi and Lou, Aaron and Purushwalkam, Senthil and Ermon, Stefano and Xiong, Caiming and Joty, Shafiq and Naik, Nikhil},
  booktitle={Proceedings of the IEEE/CVF Conference on Computer Vision and Pattern Recognition},
  pages={8228--8238},
  year={2024}
}

@article{rafailov2023direct,
  title={Direct preference optimization: Your language model is secretly a reward model},
  author={Rafailov, Rafael and Sharma, Archit and Mitchell, Eric and Manning, Christopher D and Ermon, Stefano and Finn, Chelsea},
  journal={Advances in neural information processing systems},
  volume={36},
  pages={53728--53741},
  year={2023}
}

@article{ouyang2022training,
  title={Training language models to follow instructions with human feedback},
  author={Ouyang, Long and Wu, Jeffrey and Jiang, Xu and Almeida, Diogo and Wainwright, Carroll and Mishkin, Pamela and Zhang, Chong and Agarwal, Sandhini and Slama, Katarina and Ray, Alex and others},
  journal={Advances in neural information processing systems},
  volume={35},
  pages={27730--27744},
  year={2022}
}

@inproceedings{yu2023talking,
  title={Talking head generation with probabilistic audio-to-visual diffusion priors},
  author={Yu, Zhentao and Yin, Zixin and Zhou, Deyu and Wang, Duomin and Wong, Finn and Wang, Baoyuan},
  booktitle={Proceedings of the IEEE/CVF International Conference on Computer Vision},
  pages={7645--7655},
  year={2023}
}

@misc{tan2026flowportraitreinforcementlearningaudiodriven,
      title={FlowPortrait: Reinforcement Learning for Audio-Driven Portrait Video Generation}, 
      author={Weiting Tan and Andy T. Liu and Ming Tu and Xinghua Qu and Philipp Koehn and Lu Lu},
      year={2026},
      eprint={2603.00159},
      archivePrefix={arXiv},
      primaryClass={cs.CV},
      url={https://arxiv.org/abs/2603.00159}, 
}

@inproceedings{liu2025videodpo,
  title={Videodpo: Omni-preference alignment for video diffusion generation},
  author={Liu, Runtao and Wu, Haoyu and Zheng, Ziqiang and Wei, Chen and He, Yingqing and Pi, Renjie and Chen, Qifeng},
  booktitle={Proceedings of the Computer Vision and Pattern Recognition Conference},
  pages={8009--8019},
  year={2025}
}

@article{wu2025densedpo,
  title={Densedpo: Fine-grained temporal preference optimization for video diffusion models},
  author={Wu, Ziyi and Kag, Anil and Skorokhodov, Ivan and Menapace, Willi and Mirzaei, Ashkan and Gilitschenski, Igor and Tulyakov, Sergey and Siarohin, Aliaksandr},
  journal={arXiv preprint arXiv:2506.03517},
  year={2025}
}

@InProceedings{Ng_2024_CVPR,
    author    = {Ng, Evonne and Romero, Javier and Bagautdinov, Timur and Bai, Shaojie and Darrell, Trevor and Kanazawa, Angjoo and Richard, Alexander},
    title     = {From Audio to Photoreal Embodiment: Synthesizing Humans in Conversations},
    booktitle = {Proceedings of the IEEE/CVF Conference on Computer Vision and Pattern Recognition (CVPR)},
    month     = {June},
    year      = {2024},
    pages     = {1001-1010}
}

@article{liu2025flow,
  title={Flow-grpo: Training flow matching models via online rl},
  author={Liu, Jie and Liu, Gongye and Liang, Jiajun and Li, Yangguang and Liu, Jiaheng and Wang, Xintao and Wan, Pengfei and Zhang, Di and Ouyang, Wanli},
  journal={arXiv preprint arXiv:2505.05470},
  year={2025}
}

@InProceedings{Ng_2023_ICCV,
    author    = {Ng, Evonne and Subramanian, Sanjay and Klein, Dan and Kanazawa, Angjoo and Darrell, Trevor and Ginosar, Shiry},
    title     = {Can Language Models Learn to Listen?},
    booktitle = {Proceedings of the IEEE/CVF International Conference on Computer Vision (ICCV)},
    month     = {October},
    year      = {2023},
    pages     = {10083-10093}
}

@misc{agrawal2025seamless,
      title={Seamless Interaction: Dyadic Audiovisual Motion Modeling and Large-Scale Dataset}, 
      author={Vasu Agrawal and Akinniyi Akinyemi and Kathryn Alvero and Morteza Behrooz and Julia Buffalini and Fabio Maria Carlucci and Joy Chen and Junming Chen and Zhang Chen and Shiyang Cheng and Praveen Chowdary and Joe Chuang and Antony D'Avirro and Jon Daly and Ning Dong and Mark Duppenthaler and Cynthia Gao and Jeff Girard and Martin Gleize and Sahir Gomez and Hongyu Gong and Srivathsan Govindarajan and Brandon Han and Sen He and Denise Hernandez and Yordan Hristov and Rongjie Huang and Hirofumi Inaguma and Somya Jain and Raj Janardhan and Qingyao Jia and Christopher Klaiber and Dejan Kovachev and Moneish Kumar and Hang Li and Yilei Li and Pavel Litvin and Wei Liu and Guangyao Ma and Jing Ma and Martin Ma and Xutai Ma and Lucas Mantovani and Sagar Miglani and Sreyas Mohan and Louis-Philippe Morency and Evonne Ng and Kam-Woh Ng and Tu Anh Nguyen and Amia Oberai and Benjamin Peloquin and Juan Pino and Jovan Popovic and Omid Poursaeed and Fabian Prada and Alice Rakotoarison and Rakesh Ranjan and Alexander Richard and Christophe Ropers and Safiyyah Saleem and Vasu Sharma and Alex Shcherbyna and Jia Shen and Jie Shen and Anastasis Stathopoulos and Anna Sun and Paden Tomasello and Tuan Tran and Arina Turkatenko and Bo Wan and Chao Wang and Jeff Wang and Mary Williamson and Carleigh Wood and Tao Xiang and Yilin Yang and Julien Yao and Chen Zhang and Jiemin Zhang and Xinyue Zhang and Jason Zheng and Pavlo Zhyzheria and Jan Zikes and Michael Zollhoefer},
      year={2025},
      eprint={2506.22554},
      archivePrefix={arXiv},
      primaryClass={cs.CV},
      url={https://arxiv.org/abs/2506.22554}, 
}

@article{li_learning_2017,
	title = {Learning a model of facial shape and expression from {4D} scans},
	volume = {36},
	issn = {0730-0301},
	url = {https://dl.acm.org/doi/10.1145/3130800.3130813},
	doi = {10.1145/3130800.3130813},
	number = {6},
	urldate = {2024-03-13},
	journal = {ACM Transactions on Graphics},
	author = {Li, Tianye and Bolkart, Timo and Black, Michael J. and Li, Hao and Romero, Javier},
	month = nov,
	year = {2017},
	pages = {194:1--194:17},
}

@article{Kingma2013,
	title = {Auto-{Encoding} {Variational} {Bayes}},
	url = {http://arxiv.org/abs/1312.6114},
	urldate = {2019-02-14},
	author = {Kingma, Diederik P and Welling, Max},
	month = dec,
	year = {2013},
	note = {arXiv: 1312.6114},
}

@article{van2017neural,
  title={Neural discrete representation learning},
  author={Van Den Oord, Aaron and Vinyals, Oriol and others},
  journal={Advances in neural information processing systems},
  volume={30},
  year={2017}
}

@misc{liu2026gdpo,
      title={GDPO: Group reward-Decoupled Normalization Policy Optimization for Multi-reward RL Optimization}, 
      author={Shih-Yang Liu and Xin Dong and Ximing Lu and Shizhe Diao and Peter Belcak and Mingjie Liu and Min-Hung Chen and Hongxu Yin and Yu-Chiang Frank Wang and Kwang-Ting Cheng and Yejin Choi and Jan Kautz and Pavlo Molchanov},
      year={2026},
      eprint={2601.05242},
      archivePrefix={arXiv},
      primaryClass={cs.CL},
      url={https://arxiv.org/abs/2601.05242}, 
}

@article{zhao2024image,
  title={Image and video tokenization with binary spherical quantization},
  author={Zhao, Yue and Xiong, Yuanjun and Kr{\"a}henb{\"u}hl, Philipp},
  journal={arXiv preprint arXiv:2406.07548},
  year={2024}
}

@inproceedings{zhou2025taming,
  title={Taming teacher forcing for masked autoregressive video generation},
  author={Zhou, Deyu and Sun, Quan and Peng, Yuang and Yan, Kun and Dong, Runpei and Wang, Duomin and Ge, Zheng and Duan, Nan and Zhang, Xiangyu},
  booktitle={Proceedings of the IEEE/CVF Conference on Computer Vision and Pattern Recognition},
  pages={7374--7384},
  year={2025}
}

@misc{xu2025qwen25omnitechnicalreport,
      title={Qwen2.5-Omni Technical Report}, 
      author={Jin Xu and Zhifang Guo and Jinzheng He and Hangrui Hu and Ting He and Shuai Bai and Keqin Chen and Jialin Wang and Yang Fan and Kai Dang and Bin Zhang and Xiong Wang and Yunfei Chu and Junyang Lin},
      year={2025},
      eprint={2503.20215},
      archivePrefix={arXiv},
      primaryClass={cs.CL},
      url={https://arxiv.org/abs/2503.20215}, 
}

@article{savitzky1964smoothing,
  title={Smoothing and differentiation of data by simplified least squares procedures.},
  author={Savitzky, Abraham and Golay, Marcel JE},
  journal={Analytical chemistry},
  volume={36},
  number={8},
  pages={1627--1639},
  year={1964},
  publisher={ACS Publications}
}

@inproceedings{schoneveld2025sheap,
  title={Sheap: Self-supervised head geometry predictor learned via 2d gaussians},
  author={Schoneveld, Liam and Chen, Zhe and Davoli, Davide and Tang, Jiapeng and Terazawa, Saimon and Nishino, Ko and Nie{\ss}ner, Matthias},
  booktitle={Proceedings of the IEEE/CVF International Conference on Computer Vision},
  pages={14162--14172},
  year={2025}
}

@inproceedings{zhou2022responsive,
  title={Responsive listening head generation: a benchmark dataset and baseline},
  author={Zhou, Mohan and Bai, Yalong and Zhang, Wei and Yao, Ting and Zhao, Tiejun and Mei, Tao},
  booktitle={European conference on computer vision},
  pages={124--142},
  year={2022},
  organization={Springer}
}

@article{su2024roformer,
  title={Roformer: Enhanced transformer with rotary position embedding},
  author={Su, Jianlin and Ahmed, Murtadha and Lu, Yu and Pan, Shengfeng and Bo, Wen and Liu, Yunfeng},
  journal={Neurocomputing},
  volume={568},
  pages={127063},
  year={2024},
  publisher={Elsevier}
}

@inproceedings{zhu2025infp,
  title={INFP: Audio-driven interactive head generation in dyadic conversations},
  author={Zhu, Yongming and Zhang, Longhao and Rong, Zhengkun and Hu, Tianshu and Liang, Shuang and Ge, Zhipeng},
  booktitle={Proceedings of the IEEE/CVF Conference on Computer Vision and Pattern Recognition},
  pages={10667--10677},
  year={2025}
}

@inproceedings{liu2024customlistener,
  title={Customlistener: Text-guided responsive interaction for user-friendly listening head generation},
  author={Liu, Xi and Guo, Ying and Zhen, Cheng and Li, Tong and Ao, Yingying and Yan, Pengfei},
  booktitle={Proceedings of the IEEE/CVF Conference on Computer Vision and Pattern Recognition},
  pages={2415--2424},
  year={2024}
}

@inproceedings{fan2022faceformer,
  title={Faceformer: Speech-driven 3d facial animation with transformers},
  author={Fan, Yingruo and Lin, Zhaojiang and Saito, Jun and Wang, Wenping and Komura, Taku},
  booktitle={Proceedings of the IEEE/CVF conference on computer vision and pattern recognition},
  pages={18770--18780},
  year={2022}
}

@inproceedings{richard2021meshtalk,
  title={Meshtalk: 3d face animation from speech using cross-modality disentanglement},
  author={Richard, Alexander and Zollh{\"o}fer, Michael and Wen, Yandong and De la Torre, Fernando and Sheikh, Yaser},
  booktitle={Proceedings of the IEEE/CVF international conference on computer vision},
  pages={1173--1182},
  year={2021}
}

@inproceedings{xing2023codetalker,
  title={Codetalker: Speech-driven 3d facial animation with discrete motion prior},
  author={Xing, Jinbo and Xia, Menghan and Zhang, Yuechen and Cun, Xiaodong and Wang, Jue and Wong, Tien-Tsin},
  booktitle={Proceedings of the IEEE/CVF Conference on Computer Vision and Pattern Recognition},
  pages={12780--12790},
  year={2023}
}

\clearpage
\newpage

\setcounter{page}{1}
\setcounter{section}{0}

\renewcommand{\thesection}{\arabic{section}}

\begin{center}
    \large \textbf{Supplementary Materials for GDPO-Listener: \\
    Expressive Interactive Head Generation via Auto-Regressive Flow Matching and \\ Group reward-Decoupled Policy Optimization}
\end{center}

\begin{table}[]
\centering
\setlength\tabcolsep{4pt}
\caption{User study results with 27 participants from Prolific. Numbers (\%)
indicate the proportion of users who prefer that method over all methods.}
\begin{tabular}{l|cccc}
\hline
Methods       & Lip Sync                                & Naturalness                             & Liveliness                              & Diversity                               \\ \hline
L2L           & 1.9\%                                   & 5.6\%                                   & 0.7\%                                   & 2.2\%                                   \\
DualTalk      & 27.8\%                                  & 30.0\%                                  & 17.0\%                                  & 20.4\%                                  \\ \hline
\rule[-5pt]{0pt}{14pt}\textbf{GDPO-Listener}  & \cellcolor[HTML]{EFEFEF}\textbf{70.4\%} & \cellcolor[HTML]{EFEFEF}\textbf{64.4\%} & \cellcolor[HTML]{EFEFEF}\textbf{82.2\%} & \cellcolor[HTML]{EFEFEF}\textbf{77.4\%} \\ \hline
\end{tabular}
\label{tab:user_studies}
\end{table}
\vspace{-20pt}

\section{Evaluation Metrics Details}
To complement the main paper, we detail the precise mathematical formulations of our metrics.
Lip Vertex Error (LVE) evaluates lip synchronization, while Mean Head Distance (MHD) evaluates overall geometric fidelity. Using the FLAME topology, LVE measures the maximum L2 distance among lip vertices per frame, and MHD measures the mean L2 distance across all head vertices.
Because direct vertex errors unfairly penalize diverse but plausible listening reactions, we evaluate distribution matching via Dynamic Deviation for the upper face (FDD), jaw (JDD), and head pose (PDD). These are calculated as the absolute difference between the standard deviations of frame-to-frame feature velocities for the prediction and ground truth.
We utilize Voice Activity Detection (VAD) to strictly isolate evaluation intervals: speaking segments require active actor speech, whereas listening segments strictly require the conversational partner to be speaking while the actor remains silent.

\section{Additional Quantitative Results}

To facilitate direct comparison with prior work, Tab. \ref{tab:extra_bench} presents an extended quantitative analysis on the Seamless Interaction dataset. Fréchet Distance (FD) measures the distributional realism of the generated motions, while Paired Fréchet Distance (P-FD) assesses interactive coordination by capturing the joint distributional alignment of speaker-listener behaviors. Additionally, Mean Squared Error (MSE) calculates feature-level reconstruction accuracy, SI for Diversity (SID) quantifies the behavioral diversity of the responses, and the Residual Pearson Correlation Coefficient (rPCC) measures the temporal synchrony and linear correlation between the interacting characters. 
Our GDPO-Listener shows competitive performance over previous methods.

\begin{figure}[t!]
  \centering
  \includegraphics[height=7cm]{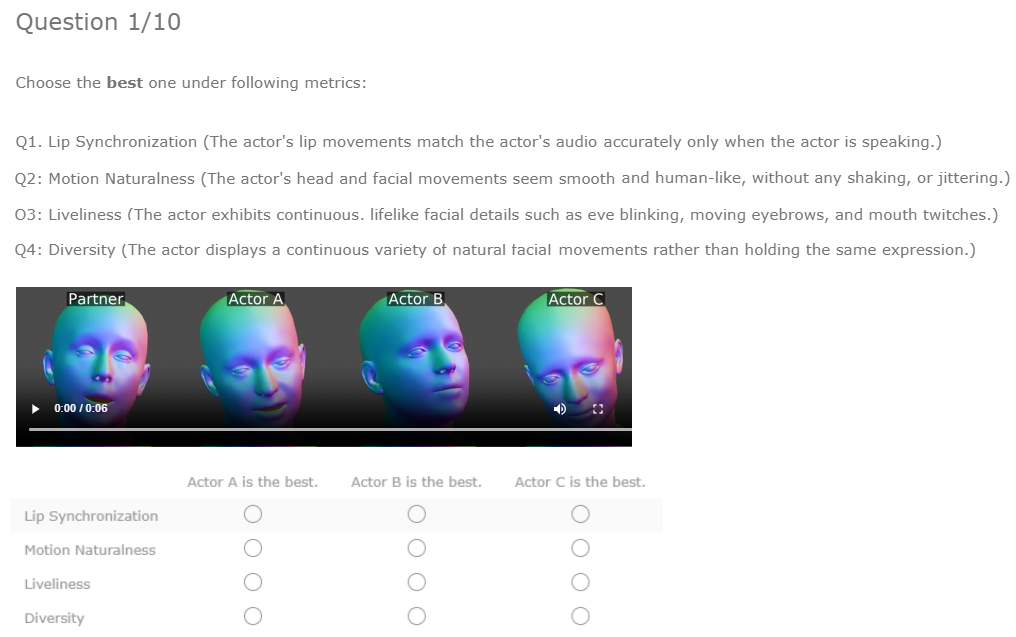}
  \caption{The interface of our user study. 
  }
  \label{fig:interface}
\end{figure}

\begin{table}[t!]
\centering
\caption{Additional Quantitative Analysis on Seamless Interaction Dataset. Following DualTalk, for better readability, FD, P-FD and MSE for pose are multiplied by 100; FD, P-FD and MSE for jaw are multiplied by 1000 and MSE for expression is multiplied by 10.}
\label{tab:extra_bench}
\resizebox{\linewidth}{!}{%
\begin{tabular}{lccc|ccc|ccc|ccc|ccc}
\hline
\multicolumn{1}{l|}{}                                               & \multicolumn{3}{c|}{FD $\downarrow$}                         & \multicolumn{3}{c|}{P-FD $\downarrow$}                       & \multicolumn{3}{c|}{MSE $\downarrow$}                       & \multicolumn{3}{c|}{SID $\uparrow$}                       & \multicolumn{3}{c}{rPCC $\downarrow$}                       \\
\multicolumn{1}{l|}{\multirow{-2}{*}{Methods}}                      & Exp            & Jaw           & Neck          & Exp            & Jaw           & Neck          & Exp           & Jaw           & Neck          & Exp           & Jaw           & Neck          & Exp            & Jaw           & Neck          \\ \hline
\multicolumn{16}{c}{\textit{Speaking}}                                                                                                                                                                                                                                                                                             \\ \hline
\multicolumn{1}{l|}{DualTalk}                                       & 93.94          & 7.49          & 8.28          & 94.92          & 7.58          & 8.44          & 19.43         & 2.89          & 3.15          & 0.93          & 1.19          & 0.87          & 0.20          & 0.21          & 0.55          \\
\rowcolor[HTML]{EFEFEF} 
\multicolumn{1}{l|}{\cellcolor[HTML]{EFEFEF}\rule[-5pt]{0pt}{14pt}\textbf{GDPO-Listener}} & \textbf{23.74} & \textbf{3.14} & \textbf{5.86} & \textbf{24.91} & \textbf{3.21} & \textbf{5.98} & \textbf{5.51} & \textbf{1.30} & \textbf{2.22} & \textbf{2.26} & \textbf{1.61} & \textbf{1.07} & \textbf{0.15} & \textbf{0.18} & \textbf{0.48} \\ \hline
\multicolumn{16}{c}{\textit{Listening}}                                                                                                                                                                                                                                                                                            \\ \hline
\multicolumn{1}{l|}{DualTalk}                                       & 94.05          & 7.16          & 9.91          & 95.06          & 7.28          & 10.08         & 19.16         & 2.63          & 3.53          & 0.69          & 0.68          & 0.48          & 0.19          & 0.28          & 0.53          \\
\rowcolor[HTML]{EFEFEF} 
\multicolumn{1}{l|}{\cellcolor[HTML]{EFEFEF}\rule[-5pt]{0pt}{14pt}\textbf{GDPO-Listener}} & \textbf{24.84} & \textbf{3.65} & \textbf{6.06} & \textbf{26.05} & \textbf{3.73} & \textbf{6.16} & \textbf{5.45} & \textbf{1.37} & \textbf{2.12} & \textbf{1.44} & \textbf{0.87} & \textbf{0.57} & \textbf{0.15} & \textbf{0.25} & \textbf{0.48} \\ \hline
\end{tabular}
} 
\end{table}

\section{User Studies}
\vspace{-5pt}

We evaluated the perceptual quality of our generated speaking and listening head motions through a user study on Prolific. 
Instead of evaluating speaking and listening in isolation, the test videos feature continuous conversational interactions to better reflect real-world performance.
A total of 27 native English speakers evaluated GDPO-Listener against DualTalk and L2L in 3-way forced-choice tasks.
We assessed four metrics: \textit{Lip Synchronization} (audio-mouth alignment), \textit{Motion Naturalness} (human-like motion realism), \textit{Liveliness} (dynamic, engaging behavior), and \textit{Diversity} (variety of generated expressions).
Fig. \ref{fig:interface} illustrates the study interface. In each task, participants watch a video featuring a reference partner alongside three anonymous generated avatars, presented in random order. Below the video, a forced-choice matrix requires users to select the best actor for each evaluation metric.
As shown in Tab. \ref{tab:user_studies}, our method is consistently preferred over all baselines across all aspects. Our method is then shown to be superior at generating realistic, dynamic conversational interactions.

\section{Video Demo}

We strongly encourage readers to view the supplementary video to fully evaluate the temporal dynamics and visual quality of our method.

\section{Extra Text Control Visualizations}

\begin{figure}[t!]
  \centering
  \includegraphics[height=6cm]{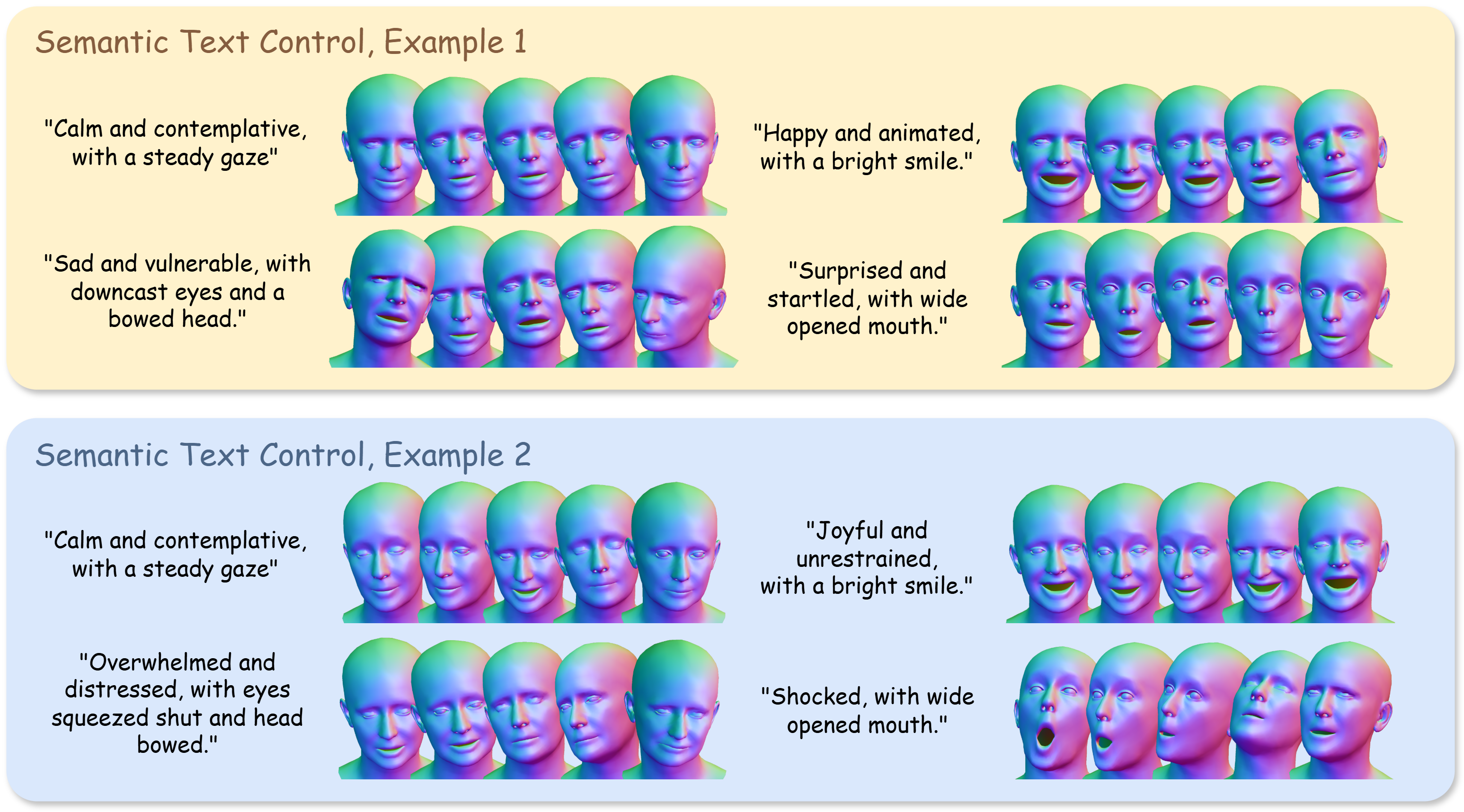}
  \caption{Additional Text Control Visualizations. 
  }
  \label{fig:extra_vis}
\end{figure}

Fig. \ref{fig:extra_vis} provides further qualitative results demonstrating our semantic text control. By conditioning on diverse prompts, GDPO-Listener accurately synthesizes the described emotional states and facial geometries. These visualizations highlight our model's capacity for fine-grained, decoupled control over conversational expressions.

\begin{table}[ht]
\centering
\caption{The explicit system prompt provided to Qwen3-VL to extract decoupled expressive text descriptions from our sampled 8-frame clips.}
\label{tab:qwen_prompt}
\begin{tabular}{p{0.95\linewidth}}
\toprule
\textbf{Qwen3-VL System Prompt} \\
\midrule
Describe the person's facial expression in a single concise sentence.  \\
1. The specific Emotion: Use precise words like `skeptical', `amused', `bored', `attentive', `confused', `happy', `angry', `calm', or `focused'. (Avoid generic `neutral'). \\
2. The specific Facial Geometry and Gaze like `narrowed eyes', `raised eyebrows', `head tilted', `looking away'. \\
\\
\textbf{Rules:} \\
1. IGNORE mouth movements related to speech. \\
2. DO NOT use words like `speaking', `talking', or `listening'. \\
3. Focus on the VIBE. Even if the face is mostly still, find the subtle emotion. \\
\\
\textbf{Examples:} \\
1. Skeptical and doubting, with one eyebrow raised and lips pursed. \\
2. Joyful and energetic, smiling broadly with crinkled eyes. \\
3. Bored and detached, looking away with a blank expression. \\
4. Attentive and focused, staring directly ahead with a slight nod. \\
\bottomrule
\end{tabular}
\end{table}
\section{Text Annotation Details}

To enable our model with text control, we require high-quality, continuous descriptions of the actors' facial behaviors. We automate this annotation process using the Qwen3-VL vision-language model. For each 200-frame video clip in our dataset, we uniformly sample 8 frames to provide the model with sufficient temporal context while maintaining computational efficiency. We specifically design the prompt to decouple the actor's underlying emotional state and facial geometry from speech-related mouth movements. The model is instructed to output a single, concise sentence capturing specific emotions and gaze dynamics while strictly ignoring active articulation. The exact system prompt, including the critical constraints and few-shot examples provided to the model, is detailed in Table \ref{tab:qwen_prompt}.

\vspace{-10pt}

\end{document}